# Region-based Convolution Neural Network Approach for Accurate Segmentation of Pelvic Radiograph

Ata Jodeiri
School of Electrical & Computer Engineering
University of Tehran
Tehran, Iran
ata.jodeiri@ut.ac.ir

Reza A. Zoroofi
School of Electrical & Computer Engineering
University of Tehran
Tehran, Iran
zoroofi@ut.ac.ir

Yuta Hiasa
Graduate School of Information Science
Nara Institute of Science and Technology
Nara, Japan
hiasa.yuta.ht7@is.naist.jp

Masaki Takao
Department of Orthopedic Surgery
Osaka University Graduate School of Medicine
Suita, Japan
masaki-tko@umin.ac.jp

Nobuhiko Sugano
Department of Orthopedic Medical Engineering
Osaka University Graduate School of Medicine
Suita, Japan
n-sugano@umin.net

Yoshinobu Sato
Graduate School of Information Science
Nara Institute of Science and Technology
Nara, Japan
yoshi@is.naist.jp

Yoshito Otake
Graduate School of Information Science
Nara Institute of Science and Technology
Nara, Japan
otake@is.naist.jp

*Abstract—* **With the increasing usage of radiograph images as a most common medical imaging system for diagnosis, treatment planning, and clinical studies, it is increasingly becoming a vital factor to use machine learning-based systems to provide reliable information for surgical pre-planning. Segmentation of pelvic bone in radiograph images is a critical preprocessing step for some applications such as automatic pose estimation and disease detection. However, the encoder-decoder style network known as U-Net has demonstrated limited results due to the challenging complexity of the pelvic shapes, especially in severe patients. In this paper, we propose a novel multi-task segmentation method based on Mask R-CNN architecture. For training, the network weights were initialized by large non-medical dataset and fine-tuned with radiograph images. Furthermore, in the training process, augmented data was generated to improve network performance. Our experiments show that Mask R-CNN utilizing multi-task learning, transfer learning, and data augmentation techniques achieve 0.96 DICE coefficient, which significantly outperforms the U-Net. Notably, for a fair comparison, the same transfer learning and data augmentation techniques have been used for U-net training.**

I. INTRODUCTION

Patient-specific pre-surgical planning plays an essential role in the success of Total Hip Arthroplasty (THA), which is one of the most prevalent orthopedic operations [1]–[3]. Implant impingement and dislocation are the most common post-operative complications that mainly happen due to a lack of accurate information about pelvic alignment [4], [5]. Recently, researchers have proposed new methods based on the artificial intelligence approach for automatic pelvic radiograph interpretation [6]. Accurate segmentation as a preprocessing block can generate more meaningful data and simplify further processing.

Traditionally, several methods such as thresholding [7], active contour model [8], and Markov random field [9] were used, which have shown to be unsatisfactory for accurate medical image segmentation. Threshold-based methods are highly depended on the selected threshold value and image's gray level histogram. Active contour models can closely detect the object edges, but the desired counter shape requires user initialization. Inappropriate initialization point or noisy condition causes poor accuracy, especially when it comes with low contrast images. The high processing cost and complexity of the Markov random field are the main weaknesses of this method, which have limited utility in dealing with a large dataset. Furthermore, manually selection of the Region of Interest (ROI) cannot be employed in dealing with the large datasets. Hence, the necessity of a fully automatic and accurate segmentation method is crucial.

In recent years, Convolutional Neural Networks (CNN) as one of the main branches of deep learning has rapidly grown in machine vision tasks, including image segmentation, object detection, and pose estimation [10]. The advantage of CNNs lies in their ability to automatically learn high-level, layered, and hierarchical abstractions from image data by end-to-end training, which led to the state of the art results. In a considerable amount of literature, U-Net [11] has been used for 2D and 3D medical image segmentation [12]–[15]. Very recently, an intuitive model, Mask Region-based Convolutional Neural Network called Mask R-CNN [16], was proposed for semantic object detection, label classification, and mask prediction which combines the Faster R-CNN [17] object detection framework with Fully Convolutional Networks (FCN) segmentation network [18].

Motivating by Mask R-CNN promising results on parallel tasks, we adopted the framework to automatically detect the ROI and segment the pelvic shape in anterior-posterior radiograph images. We also used the transfer learning paradigm to solve the small dataset problem. In this regard, firstly, whole Mask R-CNN weights were initialized by non-medical images, and then all weights were fine-tuned with our dataset. In order to generalize the trained model and prevent overfitting, some data augmentation techniques, including zooming, rotating, translating, and flipping, applied randomly.

## II. METHOD

### A. Dataset

Our dataset contains a collection of medical records, collected from 475 patients undergoing total hip replacement surgery at Osaka University Hospital. For each patient, a CT image in the supine position and anterior-posterior radiograph image in the standing position are available. We automatically segmented the pelvis in all CTs using a previously developed method [20]. Assuming a rigid transformation between supine and standing position, intensity-based 2D/3D registration was employed to align the CT image with the radiograph image. Finally, pelvic mask aligned CT was generated and manually was refined.

### B. Segmentation Network

We used a multi-task model named Mask R-CNN as a segmentation network to detect the bounding box and separate the shape of the pelvis from the background of the radiograph images. Figure 1 shows the segmentation network architecture, which consists of four parts, including data preparation, feature extraction subnetwork, region proposal subnetwork, region proposal subnetwork, and two prediction subnetworks.

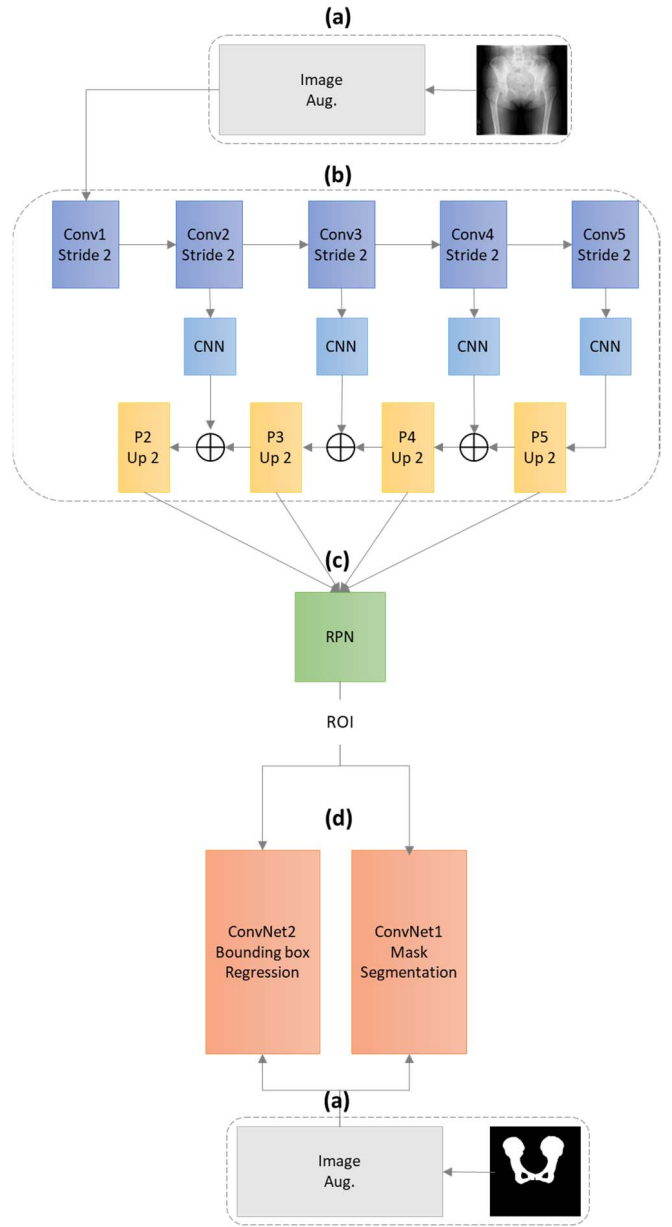

**Figure 1.** Overview of the segmentation network consisting of four parts, including **(a)** data preparation, **(b)** feature extraction subnetwork, **(c)** region proposal subnetwork, and **(d)** two prediction subnetworks.

The feature extraction block is a deep convolutional architecture for extracting the feature maps over an entire image. In this regard, Feature Pyramid Network (FPN) [19] style is employed, which consists of three pathways:

1. The bottom-up pathway made from ResNet101 extracts a collection of feature maps.

2. The top-down pathway which uses a stack of CNNs to build feature pyramid map in the same size with the bottom-up pathway.

3. The lateral connection which uses a convolution network to equalize the channel numbers and add operation

between corresponding level feature maps of bottom-up and top-down pathways.

Obtained features from the feature extraction subnetwork, are fed to the lightweight convolutional subnetwork named region proposal network (RPN). RPN roughly estimated the bounding box by checking the highest intersection over union (IoU) with pre-defined anchor boxes. Region proposal network uses 15 anchor boxes per pixel with three scales according to the input image size (1/32, 1/16, 1/8, 1/4, and 1/2) and three aspect ratios (1:1, 1:2, and 2:1) for roughly estimating the pelvic location in the 2D input image.

### C. Transfer Learning

The amount of data required for deep learning depends on the complexity of the problem and dataset variation. In dealing with medical images, the high complexity of the problem is accompanied by a small dataset. An effective technique to solve this problem is transfer learning, where the network is initially pre-trained with large non-medical images and then is fine-tuned on target dataset [29]. In this paper, the segmentation network is pre-trained on COCO dataset [21].

### D. Data Augmentation

In the training process, a random transformation was applied to the images and the corresponding masks to generalize the model capability for dealing with new images. For image transformation, random rotation in ±20º, random scaling (0.8:1.2), random horizontal and vertical translation in the scale of ±0.2, and horizontal flipping have been applied.

### III. EXPERIMENTS

#### A. Evaluation Metrics

The performance of the segmentation network is assessed by DICE coefficient. The DICE coefficient is a statistic parameter used for measuring the similarity between predicted and ground-truth segmentation by extending the spatial overlap between two binary images. The value of DICE coefficient changes from 0 for two instances with no overlap to 1 for perfect overlap. DICE coefficient for two boolean data obtained by Equation 1 ,whereas, the terms TP, FP, and FN refer to true positive, false positive and false negative, respectively.

$$DICE = \frac{2TP}{2TP + FP + FN} \quad (1)$$

#### B. Segmentation network performance

We conducted the following experiments for a detailed analysis of the performance and property of the segmentation network. In all experiments learning rate and number of epochs are set as 0.0002 and 80. Input images are resized to 256 × 256 and one image per GPU. The RPN anchor scales are set as 8, 16, 32, 64 and 128 due to input image size. In the training process, the ROI is positive if its IoU with the ground-truth box is larger than 0.7 and negative otherwise. We set all other hyperparameters following the original Mask R-CNN model. In addition to Mask R-CNN, we use three algorithmic strategies to demonstrate the effect of each bounding box regression, data augmentation and transfer learning on the proposed method. The following three scenarios were evaluated for 80 epochs, and the loss function of mask prediction on validation data has been used for comparison.

- Segmentation network without bounding box regression: we changed the multi-task system to single-task by removing the bounding box regression.

- Segmentation network without data augmentation: In the training process, all data is fed to the segmentation network without any augmenting

- Segmentation network without transfer learning: All weights initialized randomly and the network is trained from scratch.

In addition to the above two experiments, we compare the proposed method with U-Net. For U-net training, we chose the same hyperparameters as those in the original paper [20]. It is notable that the same data augmentation and transfer learning strategies have been used for U-Net training.

We divided all 475 images into three groups: 60%, 20% and 20% for training, validation, and testing respectively.

### IV. RESULTS

Figure 2 is a sample output of the Mask R-CNN model. For input radiograph image with ground-truth mask (a and b), the predicted bounding box and mask image are visualized (c and d). Figure 2 shows no significant differences between ground-truth and predicted mask images.

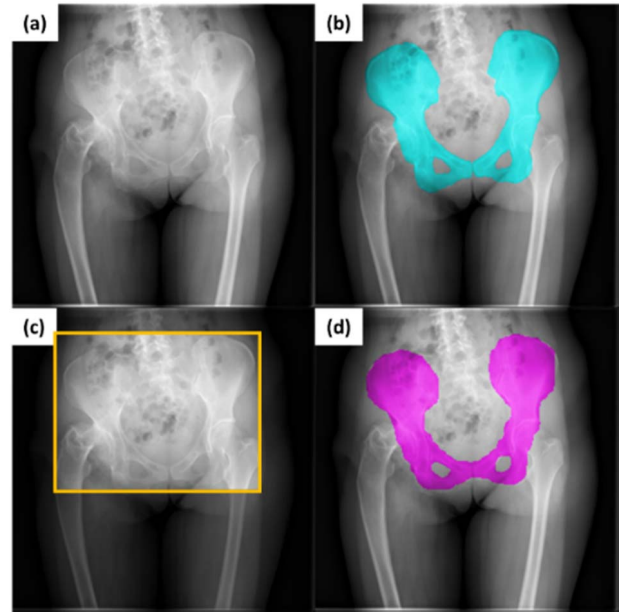

**Figure 2.** Sample output of Mask R-CNN (a) input radiograph (b) ground truth mask (c) bounding box prediction output (d) mask prediction output.

In further examinations, we will show that Mask R-CNN can sustain its high performance when applying to other cases, even severe patients.

Figure 3 shows the validation loss function of the Mask R-CNN and three scenarios, as the number of epochs increases from 1 to 80 during the training process. The results show that the loss value of all scenarios decrease rapidly in the first 30 epochs and keep reducing in three experiments with data augmentation block. The different behavior of the curve corresponding to the segmentation without data augmentation block clearly shows the model learns specific patterns of the training samples, which are irrelevant to other data and leads to higher validation loss value. Adding bounding box regression as a parallel task reduces the error at the end of all epochs from 0.095 to 0.089. Figure 3 shows that, despite the relatively large difference between the COCO dataset and radiograph images, transferring from natural images to medical images is possible and significantly effective in lowering the validation loss. The results demonstrate that the three proposed strategies, including bounding box prediction as a parallel task, data augmentation block, and transfer learning technique, all noticeably contributed to the final segmentation accuracy.

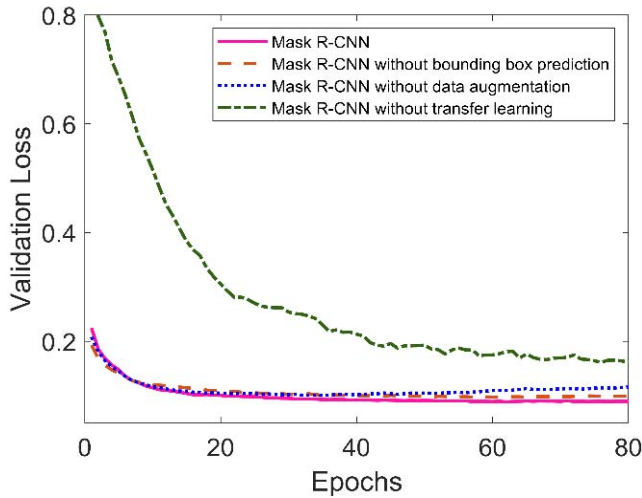

**Figure 3.** The validation loss of Mask R-CNN, Mask R-CNN without bounding box prediction, Mask R-CNN without data augmentation and Mask R-CNN without transfer learning.

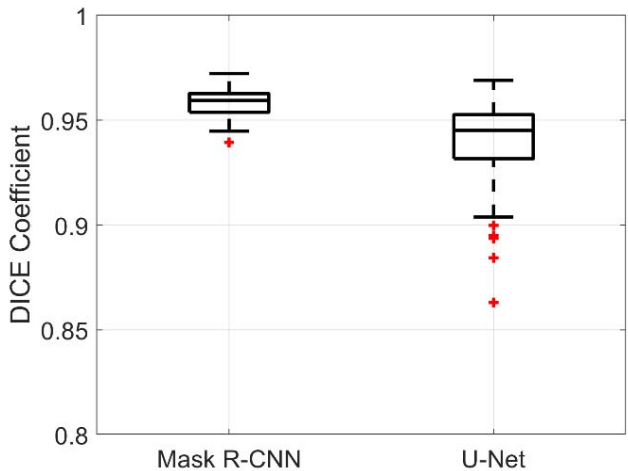

**Figure 4.** Boxplot of DICE coefficient for two segmentation networks including Mask RCNN and U-Net.

In order to evaluate the proposed model with U-Net, the box plot of DICE coefficient of Mask R-CNN and U-Net is shown in Figure 4. The box plot shows that the Mask R-CNN has consistency in segmenting all test data, while the quality of segmentation by U-Net is reduced at some samples. It can be concluded that Mask R-CNN outperforms U-Net with higher mean DICE coefficient and lower deviation.

In Figure 5, the pelvic masks estimated by Mask R-CNN and U-Net for three radiograph images are shown. By comparing the estimated masks, it can be concluded that Mask R-CNN outperforms U-Net and leads to more accurate and robust segmentation.

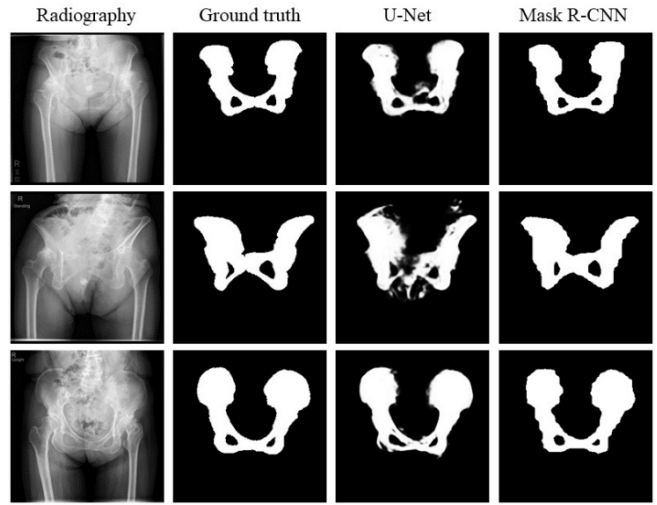

**Figure 5** Visualizing result of U-Net and Mask R-CNN for three samples from the test dataset.

In the U-Net implementation, both of the data augmentation and transfer learning strategies have been used. The role of the data augmentation block in the learning process of the U-Net is crucial, and eliminating it yield to lower segmentation accuracy. Training the U-Net, once from scratch and once from pre-train model showed us utilizing transfer learning strategy does not improve its performance in pelvic segmentation task, while transfer learning strategy is essential for Mask R-CNN due to its deeper architecture. In other words, Needing to pre-train weights is the computational burden of Mask R-CNN while the U-Net can easily train from scratch.

## V. DISCUSSION

One of the most important structural features of the Mask R-CNN that makes it an accurate model is the feature pyramid style network with ResNet101 backbone as a dominant feature map extractor. The capability of the ResNet in better learning, preventing overfitting and vanishing gradient, faster convergence speed, and better feature representation in different resolutions makes the Mask R-CNN more robust against noise and artifact. The skip connections have been proved to be effective in recovering shape structural details of the target objects. Having convolutional layers on the skip pathway of Mask R-CNN bridges the semantic gaps between encoder and decoder feature maps. This is in contrast to the plain skip connections used in U-Net, which directly connects high-resolution feature maps from the encoder to the decoder network, resulting in the fusion of semantically dissimilar feature maps. Another factor in the superiority of the Mask R-CNN is learning two related tasks simultaneously. Figure 3 showed that the proposed multi-task network yields lower validation loss. Thus, it can be concluded that adding a secondary inter-related task, i.e., pelvic bounding box prediction in the pelvic segmentation task improves the performance of the main task. In terms of the DICE coefficient, Mask R-CNN without bounding box prediction achieved 0.011 lower than Mask R-CNN but still better than U-Net.

Table I summarizes the features and performance of the U-Net and Mask R-CNN as pelvis segmentation networks.

**TABLE I.** Evaluation of the U-Net and Mask R-CNN in the pelvis segmentation task.

| Network \ Features | U-Net | Mask R-CNN |
|---|---|---|
| Network complexity | Low | High |
| Training speed | High | Low |
| Data augmentation impact | High | High |
| Transfer learning impact | Low | High |
| Bounding box prediction | No | Yes |
| ROI detection Phase | No | Yes |
| DICE coefficient | 0.9321 | 0.9584 |

## VI. Summary

In recent years, for medical image segmentation, U-net mainly has been used. Based on experiments on our task, we found that although U-net is easy to train and use but fails in challenging conditions such as severe patients or abnormal tilt angles. Most recently, Mask R-CNN achieved state-of-the-art accuracy in the semantic segmentation. In this study, we modified the Mask R-CNN to adapt to our task, as well as using transfer learning and data augmentation strategies. In addition to mask prediction output, bounding box regression as a secondary task of our multi-task model is trained.

We believe that the multi-task model, together with the framework's flexibility and accuracy, will benefit future research on radiograph image analysis.

## Acknowledgment

This research was partially supported by Japan Student Services Organization (JASSO) scholarship.

## References

[1] Z. Wan, M. Boutary, and L. D. Dorr, "The Influence of Acetabular Component Position on Wear in Total Hip Arthroplasty," *J. Arthroplasty*, vol. 23, no. 1, pp. 51–56, Jan. 2008.

[2] F. Yoshimine, "The safe-zones for combined cup and neck anteversions that fulfill the essential range of motion and their optimum combination in total hip replacements," *J. Biomech.*, vol. 39, no. 7, pp. 1315–1323, Jan. 2006.

[3] F. J. Kummer, S. Shah, S. Iyer, and P. E. DiCesare, "The effect of acetabular cup orientations on limiting hip rotation," *J. Arthroplasty*, vol. 14, no. 4, pp. 509–513, Jun. 1999.

[4] K. Uemura *et al.*, "Change in Pelvic Sagittal Inclination From Supine to Standing Position Before Hip Arthroplasty," *J. Arthroplasty*, vol. 32, no. 8, pp. 2568–2573, Aug. 2017.

[5] R. Y. Wang, W. H. Xu, X. C. Kong, L. Yang, and S. H. Yang, "Measurement of acetabular inclination and anteversion via CT generated 3D pelvic model," *BMC Musculoskelet. Disord.*, vol. 18, no. 1, pp. 1–7, Aug. 2017.

[6] A. Jodeiri *et al.*, "Estimation of Pelvic Sagital Inclanation from Anteroposterior Radiograph Using Convolutional Neural Networks: Proof-of-Concept Study," in *EPiC Series in Health Sciences*, 2018, vol. 2, pp. 114–108.

[7] S. G. Armato, M. L. Giger, and H. MacMahon, "Automated lung segmentation in digitized posteroanterior chest radiographs," *Acad. Radiol.*, vol. 5, no. 4, pp. 245–255, Apr. 1998.

[8] B. van Ginneken, A. F. Frangi, J. J. Staal, B. M. ter Haar Romeny, and M. A. Viergever, "Active shape model segmentation with optimal features," *IEEE Trans. Med. Imaging*, vol. 21, no. 8, pp. 924–933, Aug. 2002.

[9] J. Nie *et al.*, "Automated brain tumor segmentation using spatial accuracy-weighted hidden Markov Random Field," *Comput. Med. Imaging Graph.*, vol. 33, no. 6, pp. 431–441, Sep. 2009.

[10] Y. LeCun, K. Kavukcuoglu, and C. Farabet, "Convolutional networks and applications in vision," in *Proceedings of 2010 IEEE International Symposium on Circuits and Systems*, 2010, pp. 253–256.

[11] O. Ronneberger, P. Fischer, and T. Brox, "U-Net: Convolutional Networks for Biomedical Image Segmentation," in *Medical Image Computing and Computer-Assisted Intervention – MICCAI 2015*, 2015, pp. 234–241.

[12] K. Breininger, S. Albarqouni, T. Kurzendorfer, M. Pfister, M. Kowarschik, and A. Maier, "Intraoperative stent segmentation in X-ray fluoroscopy for endovascular aortic repair," *Int. J. Comput. Assist. Radiol. Surg.*, vol. 13, no. 8, pp. 1221–1231, Aug. 2018.

[13] D. Lee and H.-J. Kim, "Restoration of Full Data from Sparse Data in Low-Dose Chest Digital Tomosynthesis Using Deep Convolutional Neural Networks," *J. Digit. Imaging*, pp. 1–10, Sep. 2018.

[14] Ö. Çiçek, A. Abdulkadir, S. S. Lienkamp, T. Brox, and O. Ronneberger, "3D U-Net: Learning Dense Volumetric Segmentation from Sparse Annotation," in *Medical Image Computing and Computer-Assisted Intervention – MICCAI 2016*, 2016.

[15] J. Son, S. J. Park, and K.-H. Jung, "Towards Accurate Segmentation of Retinal Vessels and the Optic Disc in Fundoscopic Images with Generative Adversarial Networks," *J. Digit. Imaging*, pp. 1–14, Oct. 2018.

[16] K. He, G. Gkioxari, P. Dollar, and R. Girshick, "Mask R-CNN," *Proc. IEEE Int. Conf. Comput. Vis.*, vol. 2017-Octob, pp. 2980–2988, 2017.

[17] S. Ren, K. He, R. Girshick, and J. Sun, "Faster R-CNN: Towards Real-Time Object Detection with Region Proposal Networks," *IEEE Trans. Pattern Anal. Mach. Intell.*, vol. 39, no. 6, pp. 1137–1149, Jun. 2017.

[18] J. Long, E. Shelhamer, and T. Darrell, "Fully Convolutional Networks for Semantic Segmentation," in *IEEE Conference on Computer Vision and Pattern Recognition (CVPR)*, 2015.

[19] T.-Y. Lin, P. Dollár, R. Girshick, K. He, B. Hariharan, and S. Belongie, "Feature Pyramid Networks for Object Detection," in *The IEEE Conference on Computer Vision and Pattern Recognition (CVPR)*, 2017.

[20] F. Yokota, T. Okada, M. Takao, N. Sugano, Y. Tada, and Y. Sato, "Automated Segmentation of the Femur and Pelvis from 3D CT Data of Diseased Hip Using Hierarchical Statistical Shape Model of Joint Structure," in *Proc. Medical Image Computing and Computer-Assisted Intervention*, 2009, vol. 5762, pp. 811–818.

[21] T.-Y. Lin *et al.*, "Microsoft COCO: Common Objects in Context," May 2014.